\documentclass[journal]{IEEEtran_mod}

\usepackage{amsmath}
\usepackage{graphicx}
\usepackage{algorithmicx, algpseudocode}
\usepackage{float}
\usepackage{stfloats}
\usepackage{url}
\usepackage{color}
\usepackage[dvipsnames]{xcolor}
\usepackage{hyperref}

\newcommand{\bigo}{\mathcal{O}}
\newcommand{\codename}[1]{\textbf{\texttt{#1}}}
\newcommand{\kopt}{k^{\star}}
\newcommand{\kmax}{k_{\max}}
\newcommand{\slfrac}[2]{#1/#2}

\title{GALILEO: A Generalized Low-Entropy Mixture Model}
\author{\IEEEauthorblockN{\bf Cetin~Savkli,
Jeffrey~Lin, Philip~Graff, and Matthew~Kinsey}\\
\IEEEauthorblockA{JHU Applied Physics Laboratory,
11100 Johns Hopkins Road, Laurel, MD 20723, USA}
}

\begin{document}

\maketitle

\noindent\textit{\textbf{Abstract --}}
\textit{\small We present a new method of generating mixture models for data with categorical attributes.
The keys to this approach are an entropy-based density metric in categorical space and annealing of high-entropy/low-density components from an initial state with many components.
Pruning of low-density components using the entropy-based density allows \codename{GALILEO} to consistently find high-quality clusters and the same optimal number of clusters.
\codename{GALILEO} has shown promising results on a range of test datasets commonly used for categorical clustering benchmarks.
We demonstrate that the scaling of \codename{GALILEO} is linear in the number of records in the dataset, making this method suitable for very large categorical datasets.}

\vspace{1em}




\section{Introduction}\label{sec:intro}
The growth of large-scale datasets and diversity of data brings an urgency to the development of analytic methods that can handle high volume and dimensionality as well as data that include a mixture of categorical and numerical attributes.
Approaching analysis from a probabilistic perspective wherein data is represented as a high-dimensional mixture model provides a transparent representation and a tool that supports common operations such as clustering, anomaly detection, and classification.

While mixture models are a powerful tool, they are often employed for numerical data where mathematical functions, such as multivariate Gaussians, can be used.
Each mixture component concisely captures the contribution of a dense region in the high-dimensional space to the distribution as a whole.

In this paper, we present an new algorithm, the GenerALIzed Low-EntrOpy mixture model (\codename{GALILEO}), to extend mixture models to categorical attribute space using a new definition of component density that applies to categorical data.
Our concept of categorical density remediates the lack of a natural distance metric in categorical space~\cite{boriah2008similarity} and contributes to building mixture models with high-density components that represent natural clusters.

The proposed approach involves starting with a high number of initial components and using an annealing process to iteratively remove low-density components.
This procedure results in high-density/low-entropy distributions that accurately fit the data.
In each step of the process, an expectation-maximization (EM) algorithm is used to generate a fit to the data; pruning of low-density components is then performed using an entropy-based density metric.

We demonstrate that this process generates an optimal solution with respect to the density metric for the \texttt{mushroom} dataset as well as producing comparable state-of-the-art results on other datasets commonly used in the literature.

\codename{GALILEO} is easily parallelizable and scales as $\bigo (N k \log(k))$ for $N$ data points to generate a distribution with $k$ mixture components, making it suitable for use on large datasets.
Implementation and testing of the algorithm has been done on  \codename{SOCRATES}, a scalable analytics platform developed at JHU/APL~\cite{savkli2014socrates}.

This paper is organized as follows: In the next section, we introduce the concept of a generalized density metric that provides the key ingredient of the algorithm.
In Section~\ref{sec:GMM}, we present a generalization of mixture model that leverages the density metric.
Section~\ref{sec:optimalk} describes the procedure for determining the optimal number of clusters.
Then, we review similar algorithms in Section~\ref{sec:lit_rev}.
In Section~\ref{sec:results}, we present test results on various commonly used datasets.




\section{A Generalized Density Metric}\label{sec:metric}

One of the challenges in categorical space is the evaluation of the quality of a mixture component.
In numerical space, a natural measure for the quality of a component is provided by the variance of the distribution; high-variance components represent sparsely populated regions of space.

\begin{figure}[H]
\centering
\includegraphics[width=.6\columnwidth]{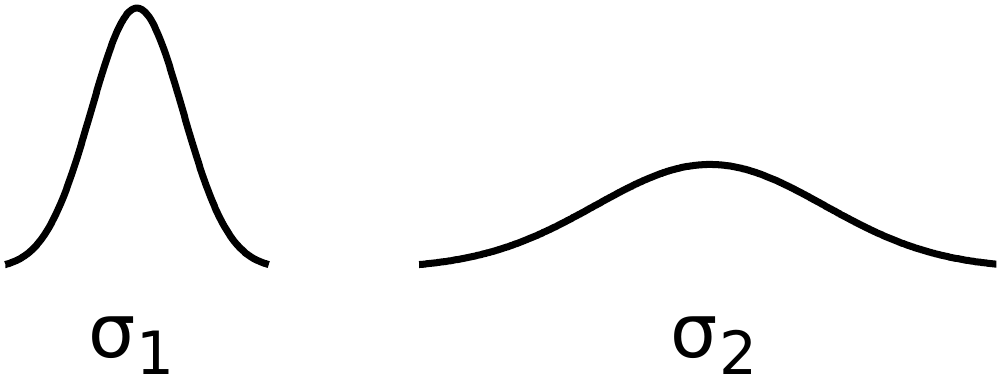}
\caption{For numerical distributions density/sparsity of a cluster can be measured concisely with variance, $\sigma^2$ (or the covariance matrix for higher dimensional data).
For example, it is evident that the density, $\rho$, $\rho_1 > \rho_2$ when $\sigma_1 < \sigma_2$ for these sample Gaussians.}
\label{fig:gaussians}
\end{figure}

When a mixture model is initialized with components far from high-density regions, the EM process steers the components towards regions with higher density to eventually find a reasonable solution.
In categorical space, the EM process is hindered by a lack of analytic representation that can leverage features of the distribution.
Combined with the lack of a component center and a universal distance metric, the EM process can lead to poor results by converging to sub-optimal distributions.
To remedy this problem, we follow an approach that starts with a high number of components in the mixture model and uses a fitness criterion and pruning process to remove low-quality components.

The fitness criterion used in pruning of low-quality components is given by a generalized density metric.
Consider, for example, the following one-dimensional distributions:
\begin{figure}[H]
\centering
\includegraphics[width=.8\columnwidth]{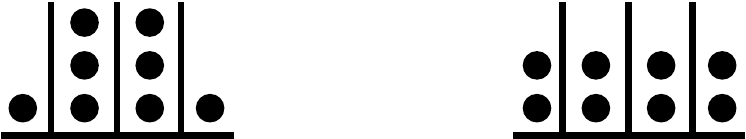}
\caption{Two distributions that span the same domain with same number of data points.
A metric similar to standard deviation is needed to describe the density of these distributions.}
\label{fig:1d_dists}
\end{figure}

Whereas a na{\"i}ve Cartesian density metric, defined as number of particles per unit length, for these two distributions is identical, the distribution on the right is clearly not as ``dense'' as distribution on the left -- i.e., the distribution on the right is more uniform than the one on the left.
We therefore propose an effective length $d$ for the axis using the entropy $S$ of the distribution,
\begin{align}
d &\equiv \exp (S), \label{eq:length} \\
S &= - \sum_{a=1}^N p_a \log p_a. \label{eq:entropy}
\end{align}
With this definition for the effective length, the length of these distributions is given by $d_L = 3.51$ and $d_R = 4$, therefore the densities ($\rho = \slfrac{N}{d}$) are $\rho_L = 2.28 > \rho_R = 2$.

As this simple example illustrates, the density definition indeed favors the left distribution.
This definition of density (Eq.~\ref{eq:length}) applies to numerical data as well as categorical data.
For example, in a Gaussian distribution, the exponentiation of the differential entropy is proportional to the standard deviation of the distribution~\cite{huber2008entropy}, i.e., $\sigma = (2 \pi e)^{-\slfrac{1}{2}} \exp (S)$.

It is possible to show that many distributions also have a similar relationship between standard deviation and entropy (e.g. $\sigma = \exp(S-1)$ for an exponential distribution and $\sigma = \exp(S)/(\sqrt{2} e)$ for a Laplace distribution).
Extending the entropy-based effective length specification to higher dimensions, the entropy-based effective volume of a hyper-cube in attribute space, with $M$ attributes, can be defined as
\begin{equation}
V = \prod_{m=1}^M d_m = \prod_{m=1}^M \exp (S_m) = \exp \left( \sum_{m=1}^M S_m \right),
\label{eq:volume}
\end{equation}
which leads to a definition of a generalized density in higher dimensions,
\begin{equation}
\rho \equiv \frac{N}{V} = N \exp \left( - \sum_{m=1}^M S_m \right).
\label{eq:density}
\end{equation}

Although it is possible to use the definition of density given in Eq.~\ref{eq:density} for both categorical and numerical variables, the numerical subspace requires some care in how entropies are defined.
If a multivariate distribution has a high degree of correlation between its variables, treating variables as independent leads to an over-estimation of the effective volume as off-diagonal regions are sparsely populated.
Therefore, it is more appropriate to define the volume of the numerical subspace in terms of entropies along the principal axes defined by Principal Component Analysis (PCA)~\cite{pearson1901principal}.
Looking to the relationship between entropy and standard deviation for guidance, the entropies of numerical subspace along principal components can be estimated using
\begin{equation}
S_{q}\equiv \log\left(\sqrt{\lambda_q}\right),
\label{eq:principalentropy}
\end{equation}
where $\lambda_q$ represent eigenvalues of the covariance matrix for numerical attributes.

\begin{figure}[H]
\centering
\includegraphics[width=.8\columnwidth]{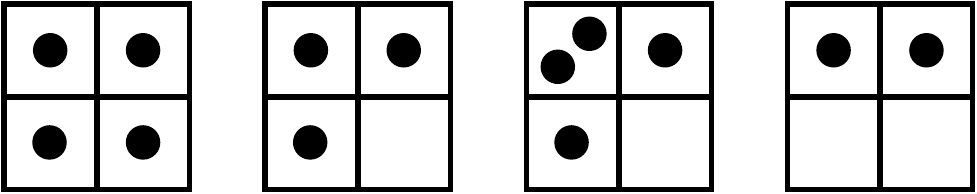}
\caption{2D distribution densities are $\rho_1=1$, $\rho_2=0.84$, $\rho_3=1.3$, $\rho_4=1$}
\label{fig:2d_dists}
\end{figure}

Some simple examples of the density calculation (Eq.~\ref{eq:density}) in two dimensions are provided by Fig.~\ref{fig:2d_dists}.

Definition of the entropy-based density implies that a uniform distribution leads to a density of $1$ independent of the size and shape of the cube for data without duplicates.
Furthermore, in this case of data without duplicate points, the density is bounded by $1$, a constraint that follows from Shannon's entropy inequality.
For a cube of $N$ particles without duplicates, the joint entropy, $S$, is given by
\begin{align}
p_a &= \frac{1}{N}, \label{eq:equalprob} \\
S &= - \sum_{a=1}^N p_a \log p_a = \log N,
\label{eq:maxentropy}
\end{align}
where $p_a$ is the probability of an individual particle.

The entropy of a multivariate distribution follows the inequality
\begin{subequations}
\begin{align}
\begin{split}
S &\leq \sum_{m=1}^M S_m
\end{split} \\
\begin{split}
\log N &\leq \sum_{m=1}^M S_m
\end{split} \\
\begin{split}
N &\leq \exp \left( \sum_{m=1}^M S_m \right)
\end{split} \\
\begin{split}
\rho &\leq 1. \label{eq:maxdensity}
\end{split}
\end{align}
\end{subequations}

Having defined a metric through Eqs.~\ref{eq:entropy},~\ref{eq:volume}, and~\ref{eq:density} to measure the quality of an individual component, we next discuss how it can be used within the context of a mixture model to generate high-density components in categorical space.




\section{Generalized Mixture Model} \label{sec:GMM}

A mixture model is defined by a superposition of probability distributions for $k$ components,
\begin{equation}
\Pr(x) = \sum_{i=1}^k \Pr(x \vert C_i) \Pr(C_i),
\end{equation}
where each component distribution, $C$, is subject to the normalization condition,
\begin{equation}
1 =  \sum_{x_j \in C} \Pr(x_j \vert C),
\end{equation}
and the components priors determine the relative size of each components,
\begin{equation}
1 =  \sum_{i=1}^k \Pr(C_i).
\end{equation}

The individual component distributions can be modeled by any suitable distribution depending on the problem and types of data attributes involved.
When the attributes are all categorical, a high-dimensional nonparametric distribution based on a clique tree may be used~\cite{savkli2016bayesian} to estimate the full joint probability.
Such a distribution is a good option when data has sub-spaces where attributes are highly correlated.
However, since individual component distributions are not required to model the entire space, but only a dense region, a complex structure such as a clique tree for individual components is not necessary.
Correlations within dense regions are much less significant and a na{\"i}ve assumption of attribute independence inside a component is typically sufficient.
The fact that a mixture model comprises many components captures the structure of correlations that a clique tree represents.
Therefore, the na{\"i}ve probability for a data point with $M$ attributes to be a member of a component is given by
\begin{equation}
\Pr(x_a|C_i) = \prod_{m=1}^{M} \Pr(x_{am} \vert C_i).
\label{eq:naivebayes}
\end{equation}
Each component contains a discrete probability distribution for each attribute of the dataset.
Numerical attributes may be considered at this time by discretizing them and treating them as categorical.
Alternatively, numerical subspaces can be represented using a multivariate distribution as is done in the Gaussian Mixture Model.
However, our focus in this paper is the generalized categorical mixture model.
The potential benefits of annealing using an entropy-based density metric for numerical data will be considered in future work.

\codename{GALILEO} starts by initializing the mixture model with a large number of components, $\kmax$.
The initialization of the components is performed by generating random component ``centers'' according to the global distribution of the data.
Since initial components need a probability distribution (and a single center point does not provide that), we use an equally-weighted average of the global distribution with the randomly generated component center.
In other words, each component starts with the probability distribution given by all $N$ data points plus a random center inserted a further $N$ times.

Following creation of the initial components, \codename{GALILEO} will then iteratively:
\begin{enumerate}
\item Use expectation-maximization to fit the distribution to data at a given $k$,
\item Sort the mixture components using the density metric (Eq.~\ref{eq:density}),
\item Prune the lowest-density components,
\end{enumerate} 
until the number of components has been reduced to $1$.
The EM algorithm in Step 1 evaluates component memberships in a probabilistic manner, assigning each data point fractionally to each component.
This fractional assignment is given by the posterior probability of a measurement belonging to a component, which follows from Bayes' theorem as
\begin{equation}
\Pr(C_i \vert x_a) = \frac{\Pr(x_a \vert C_i) \Pr(C_i)}{\Pr(x_a)}.
\label{eq:bayes}
\end{equation}
An optimal solution for $k$ is then selected using an optimality criterion as described in Section~\ref{sec:optimalk}.

A detailed description of the steps of the algorithm is provided in Fig.~\ref{fig_sim}.
\begin{figure}[h]
\centering
\begin{algorithmic}[1]
\Require $\kmax:$ Maximum component centers, $\beta:$ step root.
\Statex Initialization:
\Statex Set all component distributions $f_i$ to global distribution $G$
\State $f_i=G, i=1,\ldots,\kmax$.
\Statex Generate $\kmax$ random centers $x_i, i=1,\ldots,\kmax$ and insert each center to a distribution, $f_i$, $N$ times to define the initial mixture model:
\State $g_{k} = \left\{\left(\alpha_i, f_i\right)\right\}$ for $i=1,\ldots,\kmax$ where initially components have equal weight $\alpha_i = \slfrac{1}{\kmax}$
\State Define $k[0]=1$, $k[i+1] = k[i]+\beta^i$, $k[i_{\max}]\le \kmax$.
\State $i=i_{max}$, $\hat{k} \leftarrow k[i]$
\Statex Annealing:
\While{$\hat{k} \geq 1$}
\Statex\hspace{\algorithmicindent}Perform expectation-maximization:
\State $g_k \leftarrow EM(g_k, Data)$
\Statex\hspace{\algorithmicindent}Estimate component density \& average density:
\State $\rho_{ki} \leftarrow \rho(f_i)$ 
\State $\bar{\rho}_k \leftarrow \sum_{i=1}^k \alpha_i \rho_{ki}$
\State Use $g_k$ to calculate AIC, BIC
\Statex\hspace{\algorithmicindent}Sort components based on density:
\State $f\rightarrow {f_1,f_2, \ldots, f_{\hat{k}}}$ where $\rho_{ki}\ge\rho_{k(i+1)}$ 
\Statex\hspace{\algorithmicindent}Remove lowest density $m$ components:
\State $f\rightarrow {f_1,f_2, \ldots, f_{\hat{k}-m}}$ and $\alpha\rightarrow {\alpha_1,\alpha_2, \ldots, \alpha_{\hat{k}-m}}$ 
\Statex\hspace{\algorithmicindent}Redefine $g_k$ in terms of remaining components:
\State $g_{\hat{k}-m} = \left\{\left(\alpha_i, f_i\right)\right\}$ for $i=1,\ldots,(\hat{k}-m)$
\State $i=i-1$, $\hat{k} \leftarrow k[i]$
\EndWhile
\Statex Selection:
\State \Return Best result, $g_{\kopt}$, according to AIC, BIC, or $\bar{\rho}$
\end{algorithmic}
\caption{Algorithm to generate mixture model through density based annealing}
\label{fig_sim}
\end{figure}

This method is similar to the \textit{finite mixture model} used by~\cite{meek2002learning, cheeseman1988bayesian}, with further details in~\cite{titterington1985statistical, mclachlan1988mixture}.
However, the introduction of the denisty metric and the procedure for optimizing $k$ make \codename{GALILEO} a unique application of this model.




\section{Search for Optimal $k$}\label{sec:optimalk}

There are general rules of thumb about the relationship between the optimal number of components, $\kopt$, and the number of data points, $N$.
However, in general, it is not possible to make a definitive statement about such a relationship.
From our experiments, we find that $\kmax$ should be picked such that it is at least twice as large as the expected optimal number of components, $\kopt$.
This choice gives the annealing process the opportunity to converge to the optimal solution consistently.
Larger values of $\kmax$ will not affect the value of $\kopt$, but will take longer to converge due to more steps being required.

In practice, we use the following procedure to step down from $\kmax$.
By choosing a parameter, $\beta \geq 1$, we then inspect the set of $k$ values that are defined by the relationship,
\begin{equation}
k[i+1] = \lfloor k[i]+\beta^i \rfloor
\end{equation}
where $k[0]=1$ and $k[i_{\max}]\le \kmax$.
The parameter $\beta$ determines how finely the optimization of number of components is performed.
Using such a rule, the number of possible mixture models inspected scales as $\log(\kmax)$.

For each value of $k$, an EM procedure is performed to converge to a solution using available components at the level.
Next, the quality of the mixture model solution is measured.
Two commonly used metrics for model selection are the Akaike Information Criterion (AIC,~\cite{akaike1998information}) and Bayesian Information Criterion (BIC,~\cite{schwarz1978estimating}). These are given by
\begin{align*}
{\rm AIC} &= 2\nu - 2\log(L),\\
{\rm BIC} &= \log(N)\nu-2\log(L),
\end{align*}
where $\nu$ is the degrees of freedom of the model.
A detailed description and comparison of these metrics is given by~\cite{vrieze2012model,aho2014model}.
In addition to the AIC and BIC, we also evaluate the size-weighted average density of the components,
\begin{equation}
\bar{\rho} = \sum_{i=1}^k \alpha_i \rho_i.
\end{equation}
where 
\begin{equation}
\alpha_i = \frac{1}{N} \sum_{a=1}^N \Pr(C_i \vert x_a)
\end{equation}
and $\Pr(C_i \vert x_a)$ is given by Eqn.~\eqref{eq:bayes} with $\Pr(C_i) = \alpha_i$ and initially $\alpha_i = \slfrac{1}{\kmax}$ as we start with $\kmax$ components.
Whereas one would seek to minimize the AIC or BIC, we wish to maximize the density, $\bar{\rho}$, of the mixture model.
The density measure, $\bar{\rho}$ has the benefit over the AIC and BIC in that it scales only with the number of clusters, number of attributes, and cardinality of attributes -- there is no dependence on the dataset size so it will be simpler and faster to compute than likelihood-based metrics.
In later examples, we compare using each of these three criteria to determine $\kopt$, finding that they agree in certain cases.
The choice of which to use may be data-dependent and is up to the user to choose.




\section{Relevant Literature} \label{sec:lit_rev}

To date, most of the work in the realm of clustering algorithms has been focused on the realm of numerical data~\cite{aggarwal2013data,fahad2014survey}.
However, there has been some work done in regard to the clustering of categorical and mixed data.
In this respect, there are a handful of algorithms that represent the state of the art, namely \codename{ROCK} and \codename{COOLCAT}.
\codename{DBSCAN} is a numerical clustering algorithm that uses a density notion similar to that of \codename{GALILEO}.
In this section we will briefly review each of these algorithms.
In~\cite{Liang:2012}, the authors present a review of the clustering literature and propose a different entropy-based method for determining optimal cluserting of mixed data.
Due to space constraints, further comparisons with other algorithms are deferred to future work.

\subsection{\codename{ROCK}}\label{sec:rock}
\codename{ROCK}~\cite{guha2000rock} is often used a benchmark for the quality of a categorical clustering algorithm.
\codename{ROCK} first computes the Jaccard coefficient between all pairs of data points.
By then applying a threshold, $\theta$, to these coefficients, \codename{ROCK} assigns each data point a list of ``neighbors'' and computes the matrix $L_{ab}$, the number of common neighbors shared by points $a$ and $b$.
\codename{ROCK} then agglomeratively finds $k$ clusters that maximize the criterion function,
\begin{equation}
E_l = \sum_{i=1}^k n_i \sum_{a,b \in C_i} \frac{L(a,b)}{n_i^{1+2f(\theta)}},
\end{equation}
where $f(\theta)$ is a cluster fitness function chosen by the user that depends on the data and type of cluster desired.
While \codename{ROCK} has been shown to produce high-quality results, it suffers from a poor worst-case complexity of $\bigo (N^2 \log N)$.
Additionally, it requires the user to tune the algorithm to the data through the choice of both the thresholding parameter $\theta$ and the fitness function $f(\theta)$.

\subsection{\codename{COOLCAT}} \label{sec:coolcat}
\codename{COOLCAT}~\cite{barbara2002coolcat} uses the notion of entropy as the means to cluster the data.
The algorithm begins by selecting $k$ samples that collectively have the highest entropy.
These $k$ points will be the initial $k$ cluster centers.
\codename{COOLCAT} then proceeds by adding each sample in the dataset to the cluster that will result in the smallest increase in entropy.

As a result of this sequential process, \codename{COOLCAT} is sensitive to the ordering of the data.
In order to limit this sensitivity, the data is processed in batches and a re-clustering procedure is performed after each batch.
This procedure takes some fraction of the most poorly-fit data points and reassigns them to the clusters.

Even with the re-clustering procedure, \codename{COOLCAT} results are strongly dependent on the ordering of the data.
Moreover, the process of choosing the initial $k$ clusters is $\bigo (S^2)$ where $S$ is some representative sample of $N$, limiting \codename{COOLCAT}'s effectiveness for large datasets.

\subsection{\codename{DBSCAN}}\label{sec:dbscan}
\codename{DBSCAN}~\cite{ester1996density}, much like \codename{GALILEO}, uses a notion of density in order to find clusters of points.
Unlike the methods covered to this point, \codename{DBSCAN} is strictly for use on numerical data, requiring a distance metric to calculate distances between points in the data.
The authors have even extended \codename{DBSCAN} to cluster spatially extended objects like polygons~\cite{sander1998density}.
The algorithm is able to automatically find the number of clusters as well as find clusters of arbitrary shape.




\section{Results} \label{sec:results}

In this section we will present the results of \codename{GALILEO}'s clustering on a few publicly available datasets.
\codename{GALILEO} clusters by assigning each data point to its most probable component in the $\kopt$ components in the optimal mixture model, $g_{\kopt}$.
We first describe the datasets to be used and then demonstrate \codename{GALILEO}'s performance, including some comparisons to other algorithms mentioned previously.

\subsection{Experimental Datasets}

\subsubsection{Congressional Votes} \label{dataset:votes}
The Congressional votes dataset\footnote{\url{https://archive.ics.uci.edu/ml/datasets/Congressional+Voting+Records}}, \texttt{votes}, is from the UCI Machine Learning Repository~\cite{Lichman:2013}.
This dataset consists of the 1984 voting history of each member of Congress with respect to $16$ different issues.
Each member of Congress is assigned $16$ binary (yes/no) vote attributes as well as a classification label (Republican or Democrat).
The dataset contains $267$ Democrats and $168$ Republicans.
The classification label was ignored for the purpose of clustering so that it could be used as an independent measure of clustering results.

\subsubsection{Mushrooms} \label{dataset:mushroom}
We have also benchmarked our code using the \texttt{mushroom} dataset\footnote{\url{https://archive.ics.uci.edu/ml/datasets/Mushroom}} from the UCI Repository~\cite{Lichman:2013}.
This dataset contains the physical properties of $8124$ gilled mushrooms from $23$ species in the Agaricus and Lepiota family, as well as their edibility.
In addition to the binary edibility, there are $22$ other categorical attributes, each admitting up to twelve possible values.
These attributes describe various properties such as color, odor, and shape.
All attributes were used for clustering in order to be consistent with the procedure of the \codename{ROCK} paper~\cite{guha2000rock}.

\subsubsection{Soybean}
Another standard categorical dataset, \texttt{soybean}\footnote{\url{https://archive.ics.uci.edu/ml/datasets/Soybean+(Large)}}, consists of $19$ classes each with $35$ categorical attributes.
This dataset categorizes the properties of various types of diseases in soybeans~\cite{michalski1980learning}.
It was also obtained from the UCI Machine Learning Repository~\cite{Lichman:2013}.

\subsubsection{Zoo}
The \texttt{zoo} dataset\footnote{\url{https://archive.ics.uci.edu/ml/datasets/Zoo}} consists of $17$ different attributes related to each of $101$ species of animal.
These attributes represent, for example, how many legs an animal has or if it has feathers.
This dataset was also obtained from the UCI Machine Learning Repository~\cite{Lichman:2013}.

\subsubsection{Synthetic} \label{dataset:syn}
In order to test data of various sizes, we used \texttt{datgen}~\cite{melli1999datgen} to generate categorical datasets of arbitrary size.
These datasets were generated using a set of rules to cluster records in the attribute space.
In our tests, each record had $10$ attributes with $20$ possible values constrained by one of five rules.

\subsection{Experimental Results}

\begin{figure}[h]
\centering
\includegraphics[width=.85\columnwidth]{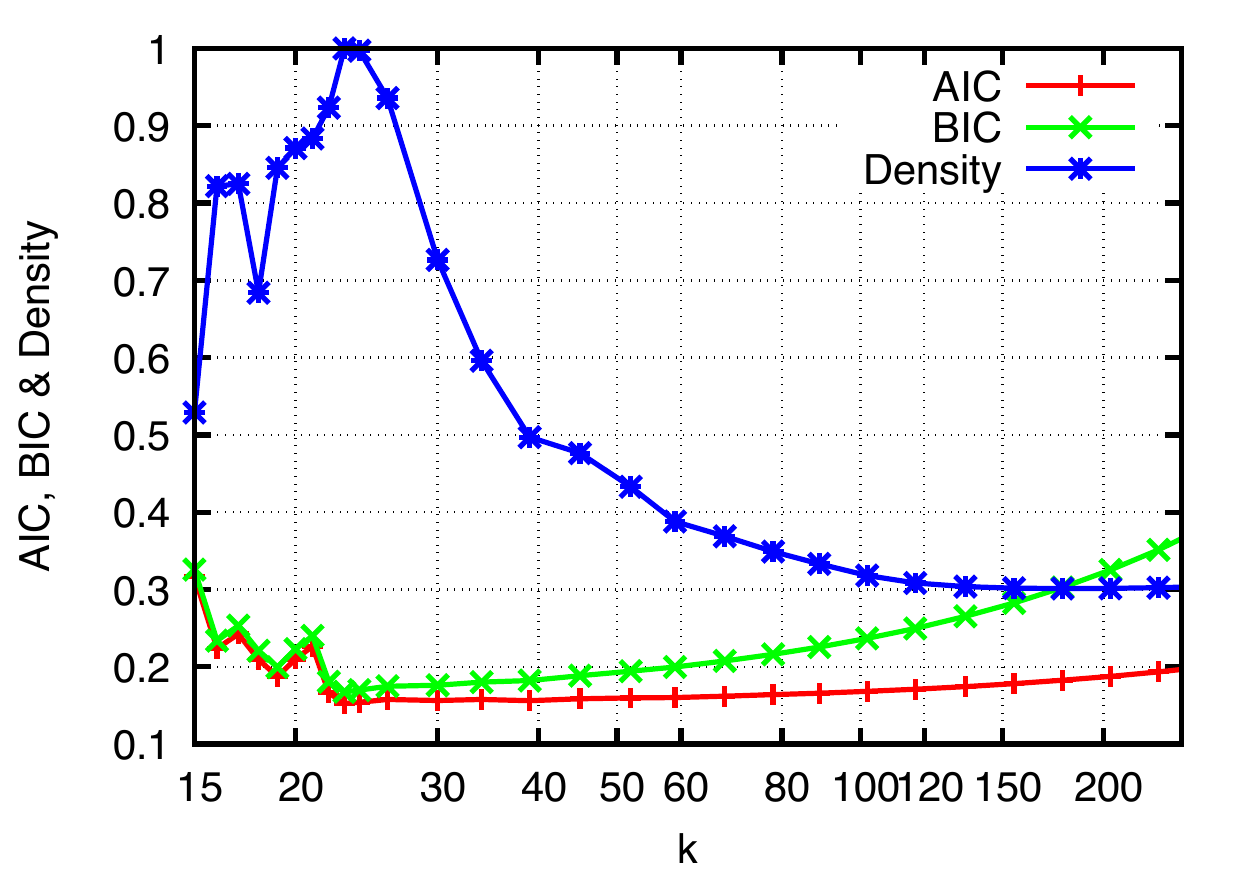}
\caption{For the \texttt{mushrooms} dataset, the optimum of all three metrics (AIC, BIC, density) coincide at $\kopt=23$ which corresponds to $\bar{\rho}=1$.}
\label{fig:aic_bic_density}
\end{figure}

In order to show in detail how the algorithm works, we use the \texttt{mushroom} dataset (Section~\ref{dataset:mushroom}).
Fig.~\ref{fig:aic_bic_density} shows the AIC, BIC, and density curves produced by \codename{GALILEO} when clustering this dataset.
All three metrics agree that $\kopt=23$, although there are visible differences in how clear this selection is.
It is worth noting that $\bar{\rho} = 1$ for $\kopt=23$.
Recall that for data that has no duplicates, the theoretical upper limit on the density of a cluster according to the inequality given by Eq.~\ref{eq:maxdensity} is $1$.
Interestingly, this optimal solution corresponds to clusters that contain only all edible or all poisonous mushrooms. Furthermore, the $23$ clusters corresponds exactly with the number of species of mushrooms represented in the dataset (unfortunately, the species identification is not in the dataset so we are unable to perform a direct comparison).
Reaching the maximum average density of $1$ in a generic clustering problem when data is categorical is clearly not always achievable.

\begin{figure}[h]
\centering
\includegraphics[width=.8\columnwidth]{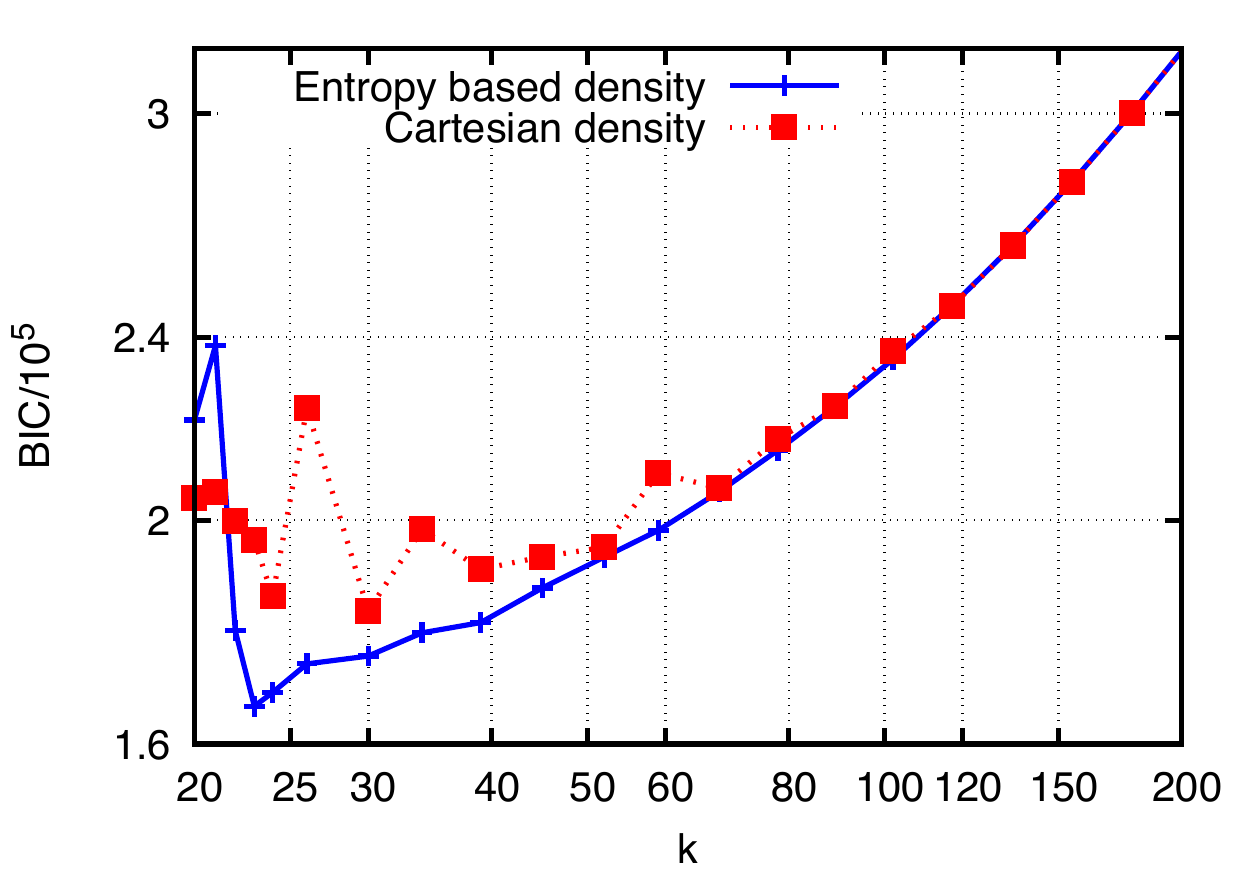}
\caption{Entropy-based density versus Cartesian density for the \texttt{mushroom} dataset.
Use of Cartesian density as a metric for pruning weak clusters does not produce an optimal solution.
The entropy-based pruning criterion leads to a solution where $\bar{\rho}=1$.}
\label{fig:cartesian_density}
\end{figure}

The role of density in obtaining this result can be understood by changing the pruning criteria from our entropy-based density to a na{\"i}ve Cartesian density.
Fig.~\ref{fig:cartesian_density} demonstrates that when a simplistic Cartesian density is used it is not possible to reach an optimal result, instead finding that $\kopt = 30$.
Whereas the results are comparable for high values of $k$, the Cartesian density is less able to determine which clusters are best to prune as the number of clusters begins to approach $\kopt$.

\begin{figure}[h]
\centering
\includegraphics[width=.85\columnwidth]{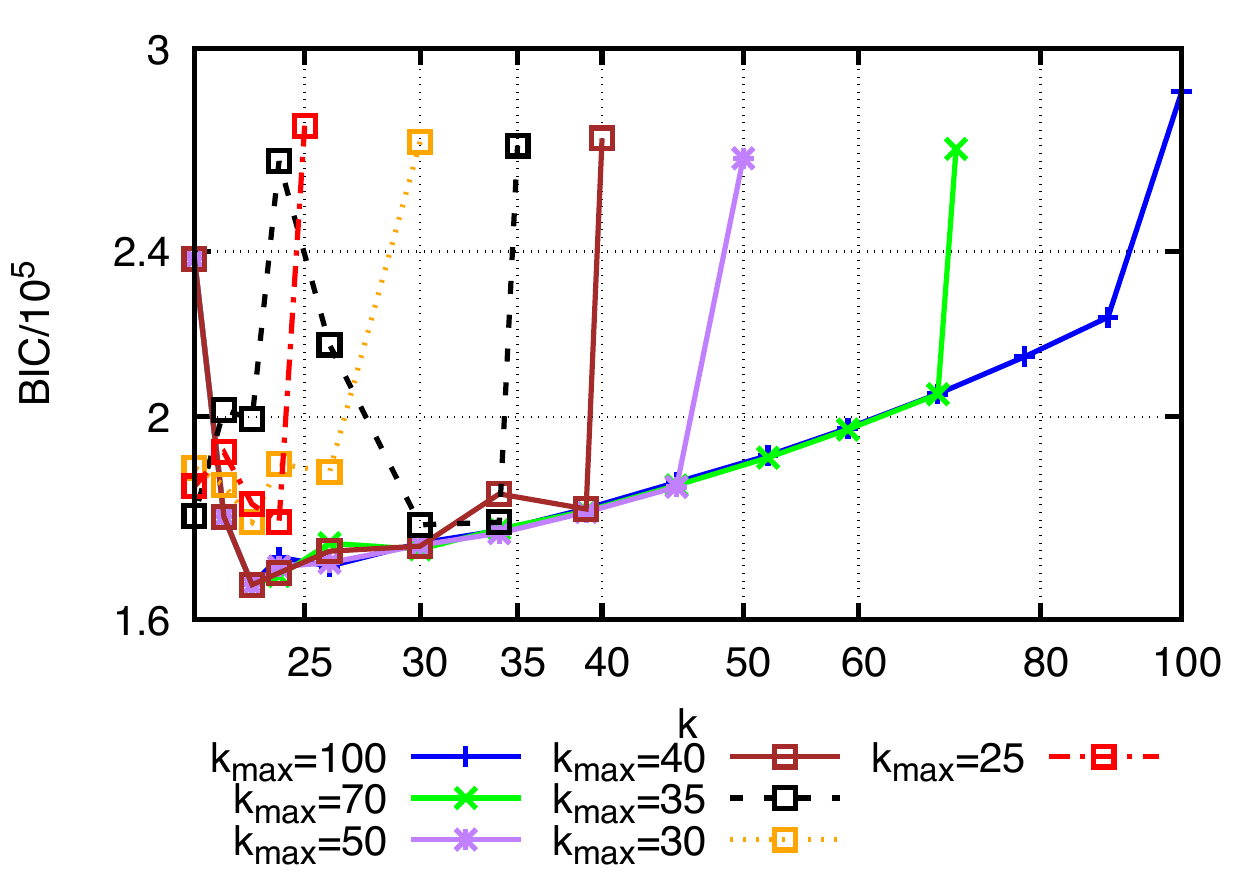}
\caption{Dependence of reaching global minimum in initial cluster size, $\kmax$, for the \texttt{mushroom} dataset.
Runs where $\kmax < 40$ do not reach the optimal solution of $\kopt=23$ clusters with $\bar{\rho}=1$.}
\label{fig:maxk}
\end{figure}

Another consideration in the execution of the algorithm is the choice of $\kmax$.
The results shown in Fig.~\ref{fig:maxk} illustrate that as long as $\kmax \geq 40$, the annealing process converges to the same optimal result, $\kopt = 23$.
If $\kmax$ is set lower, the annealing process does not have sufficient time to converge to the optimal solution.
We observe a similar behavior on other datasets tested and in general find that using a starting point that has at least twice the expected number of clusters is a good rule of thumb to reach an optimal solution.

Results represented by Figs.~\ref{fig:cartesian_density} and~\ref{fig:maxk} show that both annealing and using an entropy-based density metric contribute to achieving an optimal result.  
 
\subsection{Comparison to Other Algorithms}

We now compare \codename{GALILEO}'s results to that of other commonly-used categorical clustering algorithms (Sec.~\ref{sec:lit_rev}).
In evaluating the quality of our clustering results we will make use of the Category Utility function~\cite{categoryutility,corter1992explaining,Fisher1987}, $\bar{CU}$.
This function provides a measure of the predictive advantage gained with knowledge of the clustering relative to without that knowledge.

For a comparison of the clustering results of \codename{GALILEO} to that of \codename{ROCK}, see Tables~\ref{table:votes} and~\ref{table:mushroom}.
For the \texttt{votes} dataset, we find our results to be consistent with the clusters found using \codename{ROCK}.
Since \codename{ROCK} implements an outlier removal scheme, they report fewer total members for each cluster than we do.
While this comparison is not exact, we have, however, demonstrated an improvement in the clustering when compared to the traditional centroid-based hierarchical clustering algorithm~\cite{duda1973pattern,jain1988algorithms} that \codename{ROCK} used as a baseline.

\begin{table*}
\centering
\scriptsize
\caption{Results for the \texttt{votes} dataset for \codename{GALILEO}, \codename{ROCK}, and a traditional hierarchical method.}
\label{table:votes}
\begin{tabular}{|c||c|c|c||c|c|c||c|c|c|}
\hline
& \multicolumn{3}{|c||}{\codename{GALILEO}} & \multicolumn{3}{|c||}{\codename{ROCK}} & \multicolumn{3}{|c|}{Trad. Hier. Method} \\ \hline
Cluster & N & Rep. & Dem. & N & Rep. & Dem & N & Rep. & Dem\\ \hline
1 & 230 & 9   & 221 & 206 & 5   & 201 & 226 & 11 & 215 \\ \hline
2 & 205 & 159 & 46  & 166 & 144 & 22 & 209 & 157 & 52 \\ \hline
\end{tabular}
\end{table*}

On the \texttt{mushroom} dataset, \codename{GALILEO} finds roughly the same clusters as \codename{ROCK}, with the only exception being that \codename{GALILEO} naturally converges to $23$ clusters as opposed to $21$ with \codename{ROCK}~\cite{guha2000rock}.
These extra clusters result from splitting two of the \codename{ROCK} clusters, including the one with mixed edibility.
\codename{GALILEO} identifies no clusters with mixing in the edibility attribute.

\begin{table}[h]
\centering
\scriptsize
\caption{Clusters from found in the \texttt{mushroom} dataset by \codename{GALILEO} and \codename{ROCK}. Clusters that \codename{GALILEO} identified as two where \codename{ROCK} had one are marked with paired annotations.}
\label{table:mushroom}
\begin{tabular}{|c||c|c||c|c|}
\hline
& \multicolumn{2}{|c||}{\codename{GALILEO}} & \multicolumn{2}{|c|}{\codename{ROCK}} \\ \hline
Cluster & Edib. & Pois. & Edib. & Pois. \\ \hline
1  &   96 &    0 &   96 &    0 \\ \hline
2  &    0 &  256 &    0 &  256 \\ \hline
3$\dagger$      &  512 &    0 &  704 &    0 \\ \hline
4  &   96 &    0 &   96 &    0 \\ \hline
5  &  768 &    0 &  768 &    0 \\ \hline
6  &    0 &  192 &    0 &  192 \\ \hline
7  & 1728 &    0 & 1728 &    0 \\ \hline
8  &    0 &   32 &    0 &   32 \\ \hline
9  &    0 & 1296 &    0 & 1296 \\ \hline
10 &    0 &    8 &    0 &    8 \\ \hline
11 &   48 &    0 &   48 &    0 \\ \hline
12 &   48 &    0 &   48 &    0 \\ \hline
13 &    0 &  288 &    0 &  288 \\ \hline
14 &  192 &    0 &  192 &    0 \\ \hline
15$\ddagger$      &    0 &   72 &   32 &   72 \\ \hline
16 &    0 & 1728 &    0 & 1728 \\ \hline
17 &  288 &    0 &  288 &    0 \\ \hline
18 &    0 &    8 &    0 &    8 \\ \hline
19 &  192 &    0 &  192 &    0 \\ \hline
20 &   16 &    0 &   16 &    0 \\ \hline
21 &    0 &   36 &    0 &   36 \\ \hline
22$\ddagger$      &   32 &    0 &    0 &    0 \\ \hline
23$\dagger$      &  192 &    0 &    0 &    0 \\ \hline
\end{tabular}
\end{table}

Finally, in Table~\ref{table:summary} we report our results for the various datasets from the UCI Machine Learning repository. It is noteworthy that these values show comparable results for $\bar{CU}$ to the results of \codename{COOLCAT} (Sec.~\ref{sec:coolcat}), obtained using the \codename{coolcat-r} package\footnote{\url{https://github.com/clbustos/coolcat-r}} (except for the \texttt{mushroom} dataset -- marked with $^\ast$ --, which we obtain from~\cite{barbara2002coolcat} and normalize by an assumed $21$ clusters).

\begin{table}[h]
\centering
\scriptsize
\caption{Summary of results for UCI datasets.}
\label{table:summary}
\begin{tabular}{|r||c||c|c||c|c|}
\hline
& $k$ & $\bar{CU}_G$ & $\bar{CU}_C$ & $\bar{\rho}$ & $\bar{S}$ \\ \hline
\texttt{mushroom} & 23 & 0.3266 & 0.3393$^\ast$ & 1 & 0.2893 \\ \hline
\texttt{votes} & 2 & 1.4686 & 1.4674 & 0.01627 & 0.5988 \\ \hline
\texttt{soybean} & 7 & 1.0912 & 0.9362 & 0.02327 & 0.5151 \\ \hline
\texttt{zoo} & 9 & 0.5711 & 0.5970 & 1.4595 & 0.1882 \\ \hline
\end{tabular}
\end{table}

\subsection{Scaling Results}

\begin{figure}[h]
\centering
\includegraphics[width=.85\columnwidth]{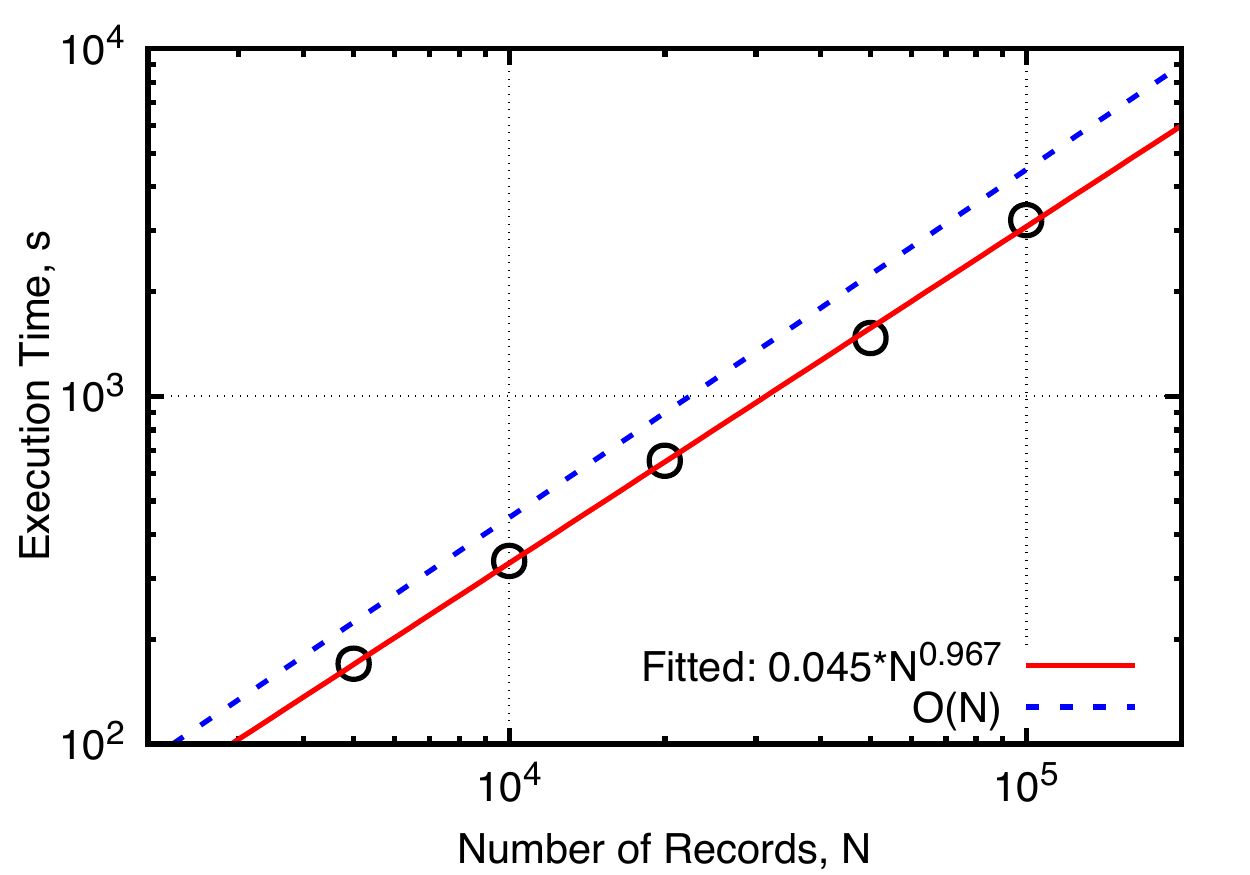}
\caption{Scaling of execution speed with respect to number of records for synthetic data. Fitting was performed for $N \geq 10^3$ to minimize the effect of overhead present at low $N$.}
\label{fig:scaling}
\end{figure}

In order to test the computational time complexity of our algorithm, we used synthetic categoric data (Section~\ref{dataset:syn}) of various sizes, built using the same rules.
For each dataset, \codename{GALILEO} was able to find the known true $\kopt$ and accurately cluster the data points.
Figure~\ref{fig:scaling} shows the timing results of this test; multiple runs were performed for each value of $N$ yielding highly consistent timings.
Once the number of records reaches a certain threshold, our scaling is very close to the theoretical time complexity $\bigo (N)$, for a fixed $k$.

\section{Conclusion}\label{sec:conclusion}

In this paper, we have presented a new method of generating mixture models in linear time for data with categorical attributes.
The keys to this approach are the entropy-based density metric in categorical space and the annealing of high-entropy/low-density components from an initial state with many components.
Pruning of low-density components using the entropy-based density allows \codename{GALILEO} to consistently find high-quality clusters and the same optimal number of clusters.
\codename{GALILEO} has shown promising results on a range of test datasets commonly used for categorical clustering benchmarks.
In particular, we have shown \codename{GALILEO}'s annealing approach and density-based pruning consistently finds the optimal clustering (based on our concept of density) on the \texttt{mushroom} dataset.
Perhaps more importantly, we have demonstrated that the scaling of \codename{GALILEO} is linear $\bigo (Nk\log(k))$ in the number of records in the dataset, making this method suitable for very large categorical datasets.
\codename{GALILEO} can be naturally extended to include numerical attributes and datasets with mixed attribute types.
In the future, we will expand the applications of this method for use on datasets consisting of mixed attributes and compare \codename{GALILEO}'s performance on numerical data to traditional numerical clustering algorithms.

\section*{Acknowledgments}
This work was supported by internal research and development funding provided by JHU/APL.

\bibliographystyle{../styles/ieee/IEEEtran}
\bibliography{biblio}

\end{document}